\documentclass[letterpaper, 10 pt, conference]{templates/ieee/ieeeconf}  

\IEEEoverridecommandlockouts                              
\usepackage{amsfonts}
\usepackage{amsmath}
\usepackage{booktabs}
\usepackage{graphicx}
\usepackage{hyperref}
\usepackage{multirow}
\usepackage{tabularx}
\usepackage{adjustbox}

\usepackage{subcaption}
\usepackage{hyperref}
\usepackage{listings}
\usepackage[style=ieee]{biblatex}
\usepackage[table,dvipsnames]{xcolor}
\usepackage{pifont}
\newcommand{\cmark}{\ding{51}}
\newcommand{\xmark}{\ding{55}}

\title{\LARGE \bf AgentGrounder: Zero-Shot 3D Visual Pointcloud Grounding using Multimodal Language Models
}

\author{
    Cuong Huynh$^{1}$, Maxim Popov$^{1}$, Denis Gridusov$^{1}$ and Sergey Kolyubin$^{1}$
    \thanks{$^{1}$ Biomechatronics and Energy-Efficient Robotics (BE2R) Lab, ITMO University, Saint Petersburg, Russia}
}

\newcommand{\name}{\textbf{AgentGrounder}}

\begin{document}
\maketitle
\thispagestyle{empty}
\pagestyle{empty}


\begin{abstract}
    3D Visual Grounding (3DVG) is an essential capability for embodied AI, requiring agents to localize objects in 3D scenes based on natural language descriptions. Recent zero-shot methods leverage 2D vision-language models (LVLMs). However, they often rely on existing sets of multi-view images and struggle with the limited semantic and spatial details provided by standard 3D segmentation tools. We present \name, a zero-shot 3D visual grounding framework that operates directly on colored point clouds without task-specific 3D training. Our approach follows a two-stage design: (1) an offline stage that applies 3D model to build an Object Lookup Table (OLT) with instance IDs, semantic labels, 3D bounding boxes; and (2) an online tool-driven agent that decomposes each query, retrieves only relevant candidates from the OLT, performs geometric scoring, and triggers image rendering on demand when additional visual evidence (e.g., color, material, or viewpoint-sensitive cues) is required. Compared with fixed anchor-target matching pipelines, this design reduces cascading matching errors and improves context-window efficiency by avoiding prompts overloaded with irrelevant objects. We evaluate on ScanRefer and Nr3D under a zero-shot setting and observe consistent improvements over SeeGround in our setup, including +2.5\% Acc@0.5 on ScanRefer and +6.3\% on Nr3D, with a notable +6.3\% gain on Nr3D view-independent queries. These results show that combining selective retrieval, geometric reasoning, and adaptive visual inspection yields a practical and robust foundation for open-vocabulary 3D grounding. Our code is available at \url{https://github.com/be2rlab/AgentGrounder}.
\end{abstract}


\section{Introduction}

3D Visual Grounding (3DVG) is a fundamental task in computer vision and robotics, aimed at localizing a target object within a 3D scene based on a natural language query. This capability is essential for enabling autonomous agents to interact seamlessly with their environment --- for instance, following a command to ``\texttt{pick up the white porcelain sink next to the counter}.'' Most existing 3DVG methods~\cite{wang2024g_g3lq, qian2024multi_mcln, huang2024chat, zheng2025video} follow a supervised learning paradigm, requiring large-scale datasets with dense annotations of 3D bounding boxes paired with linguistic descriptions. However, collecting such data is labor-intensive and often limits the model's ability to generalize to unseen object categories in open-vocabulary scenarios.

To address these limitations, zero-shot 3DVG~\cite{yang2024llm, yuan2024visual_zsvg3d, xu2025vlm, li2025seeground, yuan2024solving, zantout2025sort3d, fang2024transcrib3d} has emerged as a promising direction, leveraging the rich knowledge of Vision-Language Models (VLMs) and Large Language Models (LLMs) without task-specific 3D training. Despite encouraging progress, current zero-shot pipelines still face key issues: (1) query reasoning is often brittle due to early anchor-target matching errors, (2) visual inspection is not always selective, causing unnecessary computation and token overhead, and (3) geometric relations in 3D scenes are not consistently exploited in a transparent and deterministic manner.

In this paper, we propose \name, a zero-shot 3DVG framework that operates directly on colored 3D point clouds with a tool-driven agent. Our method follows a two-stage design. First, we run 3D model to obtain instance-level segmentation and build an Object Lookup Table (OLT) containing object IDs, semantic labels, centers, and box sizes. Second, an online LVLM agent performs query decomposition, retrieves only relevant objects from the OLT, applies deterministic geometric scoring, and triggers image rendering on demand when visual evidence (e.g., color, material, or viewpoint-sensitive cues) is required. Compared with fixed matching pipelines, this design reduces cascading matching errors and improves context efficiency by avoiding prompts filled with irrelevant objects.

Our primary contributions are summarized as follows:
\begin{itemize}
    \item We introduce a practical two-stage zero-shot 3DVG pipeline that uses only colored point clouds: offline OLT construction and online tool-driven agent reasoning.
    \item We propose a transparent grounding strategy that combines selective candidate retrieval, deterministic geometric scoring, and on-demand rendering, improving both reasoning robustness and context-window efficiency for LVLM inference.
    \item We validate the approach on ScanRefer~\cite{chen2020scanrefer} and Nr3D~\cite{achlioptas2020referit3d}, achieving consistent gains over SeeGround in our setting, including +2.5\% Acc@0.5 on ScanRefer and +1.5\% overall on Nr3D, with a notable +7.6\% on Nr3D view-independent queries.
\end{itemize}


\section{Related Work}

\textbf{3D Visual Scene Grounding}. Existing frameworks for 3D visual grounding span a continuum from fully supervised to zero-shot paradigms, each offering distinct benefits and computational trade-offs. On the supervised side, recent progress in 3D scene generation and reconstruction has yielded methods that integrate large language models (LLMs) for richer multimodal understanding and controllable scene synthesis~\cite{wang2024g_g3lq, qian2024multi_mcln, unal2024four_concretenet, huang2024chat, zheng2025video}. 

In contrast, zero-shot approaches remove the need for task-specific 3D training data by exploiting 2D foundation models and pretrained 3D backbones, using LLMs and vision–language models (VLMs) to localize objects in 3D scenes without extensive annotations or predefined taxonomies. LLM-G~\cite{yang2024llm} employs LLMs as agents that parse natural-language queries and synthesize grounding instructions for 3D environments, whereas ZSVG3D~\cite{yuan2024visual_zsvg3d} adopts a visual programming formulation with view-independent, view-dependent, and functional modules to resolve complex spatial relations through dialog-based LLM interaction. VLM-Grounder~\cite{xu2025vlm} leverages VLMs for open-vocabulary grounding, SeeGround~\cite{li2025seeground} bridges 2D VLMs and 3D scenes by rendering query-aligned images and fusing them with spatial textual descriptions via perspective adaptation and fusion alignment modules, and CSVG~\cite{yuan2024solving} addresses grounding using structured visual solving strategies. Moreover, Sort3D~\cite{zantout2025sort3d} utilizes sorting-based mechanisms for precise 3D object localization, while Transcrib3D~\cite{fang2024transcrib3d} performs scene transcription to obtain intermediate textual representations that facilitate effective grounding. 

Overall, supervised methods typically attain higher accuracy at the cost of substantial annotation effort, whereas zero-shot approaches trade some performance to enable deployment in data-scarce settings and support open-vocabulary reasoning over previously unseen object categories.

\textbf{3D Scene Segmentation}. Recent open-vocabulary 3D segmentation approaches, such as Mask3D~\cite{takmaz2023openmask3d} and ISBNet3D~\cite{ngo2023isbnet}, employ 3D mask prediction networks trained on large-scale point cloud datasets to achieve zero-shot generalization to novel semantic categories without relying on closed-set label spaces. In contrast, SAM3D~\cite{yang2023sam3d} transfers 2D foundation segmentation models, such as Segment Anything, into 3D via efficient lifting mechanisms, emphasizing interactive prompting for both semantic and instance-level tasks while exploiting vision–language model (VLM) priors. Open3DIS~\cite{nguyen2024open3dis} further extends this line of work by performing open-vocabulary 3D instance segmentation through projections of purely 2D VLM features onto point clouds, thereby obviating the need for specialized 3D networks and enhancing practicality. Any3DIS~\cite{nguyen2025any3dis} advances this paradigm by integrating foundation LLMs with VLMs for general 3D scene understanding, enabling open-vocabulary instance segmentation via multimodal prompting without task-specific 3D training. Collectively, these methods span a continuum from 3D-native models with originally closed-set assumptions (e.g., early Mask3D~\cite{takmaz2023openmask3d}) to fully open-vocabulary 2D-to-3D adaptations, illustrating a broader transition toward synergistic use of foundation models, combining VLMs, LLMs, and segmentation backbones, to realize scalable and flexible 3D scene parsing.

\textbf{Multimodality for Spatial Understanding}. Foundation models with multimodal capabilities, such as multimodal LLMs, exhibit strong generalization that is particularly desirable in 3D settings, where large-scale annotated data remain scarce. Recent studies~\cite{nguyen2024open3dis, gu2024conceptgraphs, xu2025vlm} demonstrate how 2D knowledge can be projected into 3D representations, enabling a new level of scene understanding. ConceptFusion~\cite{jatavallabhula2023conceptfusion} fuses pixel-aligned features from foundation models into 3D maps, allowing zero-shot, multimodal (text, image, audio, geometry) queries and spatial reasoning without additional training. PhyGrasp~\cite{guo2025phygrasp} combines language and 3D point-cloud representations to generalize robotic grasping via human-like physical commonsense reasoning. VLTNet~\cite{wen2025zero} introduces a vision–language reasoning module that constructs semantic maps and performs frontier-based exploration to achieve accurate language-driven zero-shot object navigation in unseen environments. The datasets proposed in~\cite{chen2020scanrefer, achlioptas2020referit3d, zhang2025iref, jia2024sceneverse} pair scenes with language descriptions to enable models to understand 3D environments, and collectively highlight that 3D scene understanding benefits from diverse annotation paradigms (synthetic versus natural language), explicit encoding of spatial relations, and robustness to linguistic variation.

\section{Methodology}

\begin{figure*}[htbp]
    \centering
    \includegraphics[width=\textwidth]{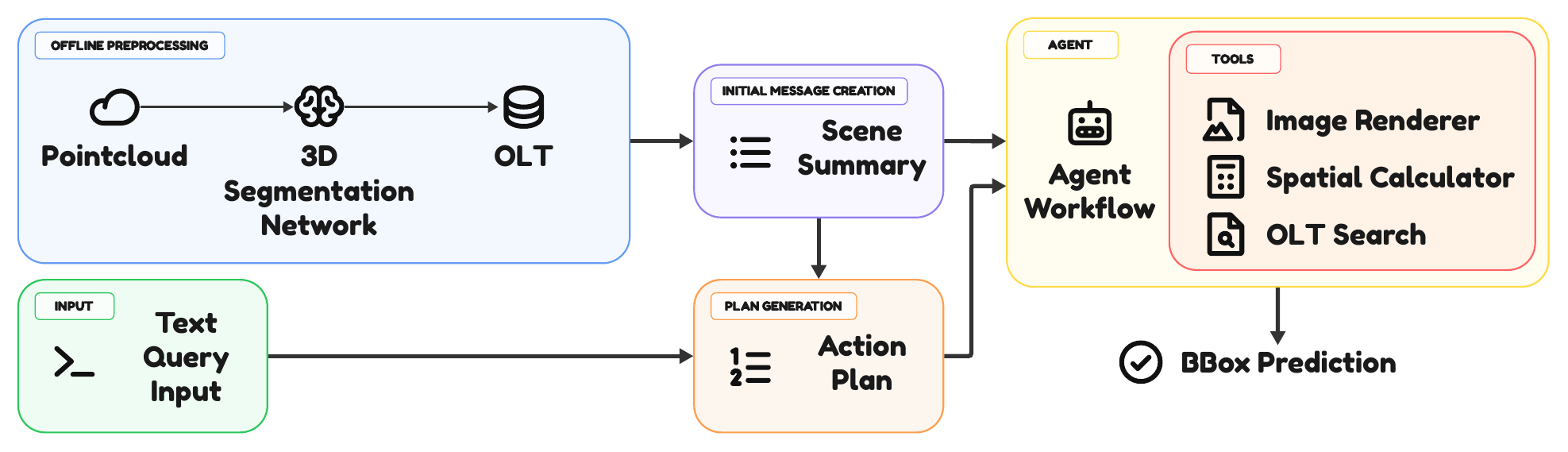}
    \caption{\textbf{Overview of the \name{} agent pipeline}. For each query, the agent first writes an explicit plan, retrieves candidate objects by semantic labels from scene metadata, and computes geometric relations (e.g., nearest/farthest, left/right, below) from 3D centers and box sizes. For view-dependent expressions, the agent calls a rendering tool to inspect selected object IDs and resolve ambiguity. Finally, it returns a structured answer containing the predicted object ID and a textual justification.}
    \label{fig:pipeline}
\end{figure*}

\textbf{Overview.} We study 3D visual grounding where, given a language query $Q$ and a scene $S$, the system predicts the target object and its 3D box. Our pipeline starts with an offline 3D instance segmentation stage that predicts object instances and their semantic descriptions from the colored point cloud. These predictions are converted into object-level metadata (ID, label, center, and box size), which is then used by the online VLM-based agent for tool orchestration and reasoning. The task is formulated as:
\begin{equation}
bbox = 3DVG(S, Q)
\end{equation}
where $bbox$ is obtained by first selecting object ID $\hat{o}$ and then retrieving its geometry from the Object Lookup Table.

The online inference pipeline is tool-driven rather than end-to-end neural prediction at query time: plan generation, candidate retrieval by label, geometric scoring, optional image-based disambiguation, and structured final output.

\subsection{Initial 3D Instance Segmentation}
As the first stage, we run a 3D instance segmentation network on scene $S$ to obtain per-object masks and initial semantic labels, following the approach in SeeGround~\cite{li2025seeground}:
\begin{equation}
\{(mask_i, sem_i)\}_{i=1}^{N} = Seg(S).
\end{equation}
For each predicted mask, we compute the corresponding 3D bounding box:
\begin{equation}
bbox_i = Bound(mask_i).
\end{equation}
This stage is executed offline once per scene and provides the object candidates used in subsequent online grounding.

\subsection{Object Lookup Table}
Following~\cite{li2025seeground}, each scene is represented as an Object Lookup Table (OLT) constructed from the segmentation outputs:
\begin{equation}
\mathcal{O}=\{(id_i, \ell_i, c_i, d_i)\}_{i=1}^{N},
\end{equation}
where $\ell_i$ is the semantic label, $c_i\in\mathbb{R}^3$ is the object center, and $d_i\in\mathbb{R}^3$ is box size. During online inference, the agent queries this table using tool calls (e.g., by label) and obtains pre-calculated 3D segmentation results.

\subsection{Agent Planning and Query Decomposition}
Given query $Q$, the agent first produces an explicit plan (tool usage order and decision logic), then extracts key semantic anchors (e.g., object categories and spatial constraints) used to form retrieval requests. This stage defines which labels and relations should be checked:
\begin{equation}
(\mathcal{L}, \mathcal{R}) = PlanExtract(Q),
\end{equation}
where $\mathcal{L}$ is the set of candidate labels and $\mathcal{R}$ contains spatial predicates (e.g., \texttt{next to}, \texttt{left of}, \texttt{below}, \texttt{closest to}).

\subsection{Candidate Retrieval and Geometric Scoring}
The agent retrieves candidates by labels from $\mathcal{O}$:
\begin{equation}
\mathcal{C} = \{o_i \in \mathcal{O} \mid \ell_i \in \mathcal{L}\}.
\end{equation}
For distance-based predicates, it computes
\begin{equation}
d(i,j) = \lVert c_i - c_j \rVert_2.
\end{equation}
For size constraints, it uses box dimensions (e.g., projected area/volume proxies). For directional predicates (left/right, above/below), it applies coordinate comparisons in the selected reference frame. A deterministic score is then used to rank candidates:
\begin{equation}
s(o_i \mid Q) = f_{geo}(o_i, \mathcal{R}, \mathcal{O}), \qquad
\hat{o} = \arg\max_{o_i \in \mathcal{C}} s(o_i \mid Q).
\end{equation}

\subsection{View-Dependent Disambiguation by Rendering}
If the query contains viewpoint-sensitive language (e.g., \texttt{when facing}, \texttt{on the right}), the agent calls a rendering tool over a subset of candidate IDs and uses visual evidence to resolve ambiguity:
\begin{equation}
I = Render(S, \mathcal{I}_{cand}), \qquad
\hat{o} = Resolve(\hat{o}, I, Q).
\end{equation}
For non-view-dependent queries with clear geometric ranking, this step is skipped.

\subsection{Robust Handling of Label Mismatch}
When user terms are absent from available labels (e.g., \texttt{board}, \texttt{fridge}), the agent maps them to the nearest available category based on context (e.g., \texttt{tv}, \texttt{kitchen cabinet}) and continues geometric reasoning. This fallback improves query coverage without retraining.

\subsection{Final Prediction}
The final output is a structured pair containing object ID and explanation:
\begin{equation}
\hat{bbox} = BBox(\hat{o}), \qquad
\text{answer}=(\hat{o}, \hat{bbox}, \text{rationale}).
\end{equation}
This formulation captures the agent behavior during inference: metadata retrieval, geometric reasoning, selective image calls for view dependence, and structured final reporting.
\section{Experiments}

\begin{figure*}[htbp]
    \centering
    \includegraphics[width=\textwidth]{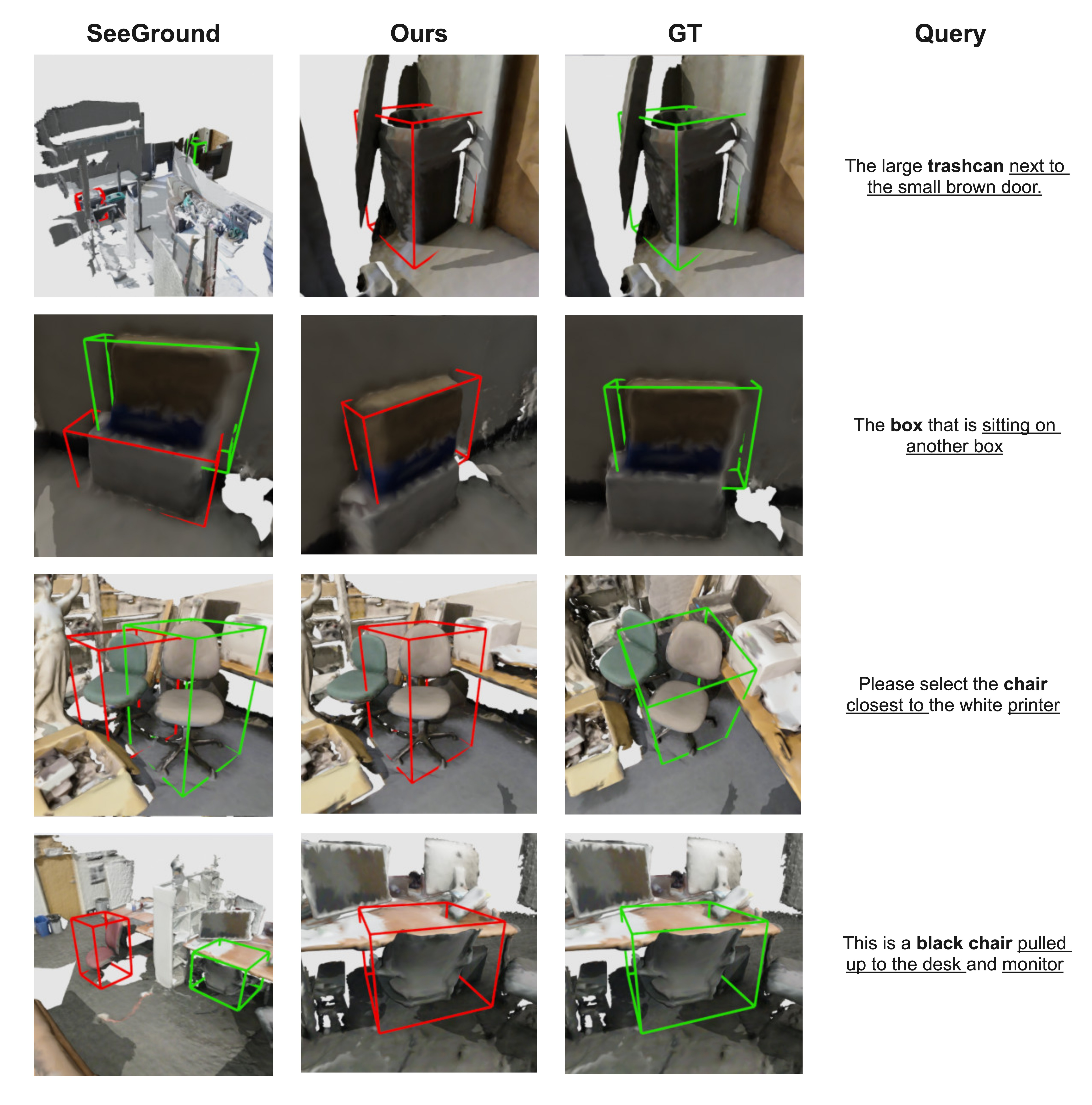}
    \caption{Results comparison between SeeGround~\cite{li2025seeground} and \name{} (Ours)} 
    \label{fig:results_comparison}
\end{figure*}

\subsection{Experimental Settings}

\textbf{Datasets.} We evaluate our proposed 3DVG approach on two widely-used benchmark datasets: ScanRefer \cite{chen2020scanrefer} and Nr3D \cite{achlioptas2020referit3d}. ScanRefer provides natural language descriptions for objects in ScanNet scenes, requiring the model to distinguish target objects based on spatial context and geometric features. Queries are categorized as \textit{Unique} or \textit{Multiple} based on the presence of same-class distractors. Nr3D, part of the ReferIt3D suite, contains descriptions collected via a reference game, classified into \textit{Easy} or \textit{Hard} levels and further divided by viewpoint dependency (\textit{View-Dependent} or \textit{View-Independent}). To ensure comprehensive evaluation, we evaluate on the full validation set of Nr3D (6,307 queries) and draw a 1,000-query subsample from ScanRefer using stratified random sampling to preserve the \textit{Unique}/\textit{Multiple} split.

\textbf{Implementation Details.} Our framework follows a two-stage pipeline. In the offline preprocessing stage, we employ Mask3D~\cite{schult2023mask3d} for 3D instance segmentation to construct an Object Lookup Table (OLT) for each scene. The OLT stores object-level metadata, including instance ID, semantic label, object center, and 3D bounding box size. 

For the online reasoning stage, we use Qwen3-VL-32B-Instruct~\cite{yang2025qwen3} as the core Vision-Language Model (LVLM). The model is deployed locally using the Ollama framework~\cite{marcondes2025using} and interfaced through the LangChain API~\cite{chase2022langchain}. Given a query, the agent invokes tools to retrieve relevant candidates from the OLT, perform geometric reasoning, and call image rendering whenever additional visual evidence is required (e.g., color, surface texture, or viewpoint-dependent cues). Compared with SeeGround~\cite{li2025seeground}, which processes queries by separating anchor and target objects and then applies NLP matching to decide what to render, our pipeline allows the agent to autonomously decide which list of objects should be rendered at runtime. This design reduces matching-induced errors and keeps rendered images focused on query-relevant objects. We follow the evaluation protocol established in ZSVG3D \cite{yuan2024visual_zsvg3d}, using accuracy with IoU thresholds between predicted and ground-truth 3D bounding boxes as the primary metric, and report results on a representative subset of the test set. All corner-case analyses are performed on the Nr3D dataset. Experiments are performed on a workstation equipped with two NVIDIA RTX 4090 GPUs, each providing 24 GB of VRAM.




\begin{table*}[t]
    \centering
    \caption{Evaluation results of 3DVG methods on \textit{ScanRefer}~\cite{chen2020scanrefer} validation set. Results are reported for \textit{``Unique"} (scenes with a single target object) and \textit{``Multiple"} (scenes with distractors of the same class) subsets, along with overall performance.}
    \vspace{-0.2cm}
    \resizebox{\linewidth}{!}{
    \begin{tabular}{r|c|c|c|cc|cc|cc}
        \toprule 
        \multirow{2}{*}{\textbf{Method}} & \multirow{2}{*}{\textbf{Venue}} & \multirow{2}{*}{\textbf{Zero-Shot}} & \multirow{2}{*}{\textbf{Agent}} & \multicolumn{2}{c|}{\textbf{Unique}} & \multicolumn{2}{c|}{\textbf{Multiple}} & \multicolumn{2}{c}{\textbf{Overall}} 
        \\
        & & & & \textbf{Acc@$\mathbf{0.25}$} & \textbf{Acc@$\mathbf{0.5}$} & \textbf{Acc@$\mathbf{0.25}$} & \textbf{Acc@$\mathbf{0.5}$} & \textbf{Acc@$\mathbf{0.25}$} & \textbf{Acc@$\mathbf{0.5}$} 
        \\
        \midrule\midrule
        G3-LQ \cite{wang2024g_g3lq}                     & CVPR'24 & \xmark & - & $88.6$ & $73.3$ & $50.2$ & $39.7$  & $56.0$ &   $44.7$  \\
        MCLN \cite{qian2024multi_mcln}                  & ECCV'24 & \xmark & - & $86.9$ & $72.7$ & $52.0$ & $40.8$  & $57.2$ &   $45.7$ \\
        ConcreteNet \cite{unal2024four_concretenet}     & ECCV'24 & \xmark & - & $86.4$ & $82.1$ & $42.4$ & $38.4$  & $50.6$ &   $46.5$ \\
        Chat-Scene \cite{huang2024chat}                 & NIPS'24 & \xmark & Vicuna-7B      & $89.6$ & $82.5$ & $47.8$ & $42.9$  & $55.5$ &   $50.2$ \\
        Video-3D LLM \cite{zheng2025video}              & CVPR'25 & \xmark & LLaVA-Video 7B & $88.0$ & $78.3$ & $50.9$ & $45.3$  & $58.1$ &   $51.7$ \\
        GPT4Scene \cite{qi2025gpt4scene}                & ICLR'26 & \xmark & Qwen2-VL-7B    & $90.3$ & $83.7$ & $56.4$ & $50.9$  & $62.6$ &   $57.0$ \\
        \midrule
        LLM-G \cite{yang2024llm}                        & ICRA'24 & \cmark & GPT-3.5            & - & -  & - & - & $14.3$ & $4.7$  \\
        LLM-G \cite{yang2024llm}                        & ICRA'24 & \cmark & GPT-4 turbo        & - & -  & - & - & $17.1$ & $5.3$  \\
        ZSVG3D \cite{yuan2024visual_zsvg3d}             & CVPR'24 & \cmark & GPT-4 turbo        & $63.8$ & $58.4$ & $27.7$ & $24.6$ & $36.4$ & $32.7$ \\
        SeeGround \cite{li2025seeground}                & CVPR'25 & \cmark & Qwen2-VL-72B       & $\underline{75.7}$ & $\underline{68.9}$ & $34.0$ & $\underline{30.0}$ & $44.1$ & $39.4$ \\
        CSVG \cite{yuan2024solving}                     & BMVC'25 & \cmark & Mistral-Large-240B  & $68.8$ & $61.1$ & $\textbf{38.4}$ & $27.3$ & $\textbf{49.6}$ & $\underline{39.8}$ \\
        \name{} (Ours)                                    & -       & \cmark & Qwen3-VL-32B       & $\mathbf{80.3}$ & $\mathbf{73.7}$ & $\underline{36.6}$ & $\textbf{31.7}$ & $\underline{47.2}$ & $\mathbf{41.9}$ \\
        \bottomrule
    \end{tabular}
    }
    \label{tab:scanrefer_results}
\end{table*}
\begin{table}[t]
    \centering
    \caption{Performance results on \textit{Nr3D}~\cite{achlioptas2020referit3d} validation set. Queries are labeled as \textit{``Easy"} (only one distractor) or \textit{``Hard"} (with multiple distractors), and as \textit{``View-Dependent"} or \textit{``View-Independent"} based on viewpoint requirements for grounding.}
    \begin{tabular}{r|cccc|c}
        \toprule
        \textbf{Method} & \textbf{Easy} & \textbf{Hard} & \textbf{Dep.} & \textbf{Indep.} & \textbf{Overall} 
        \\
        \midrule\midrule
        \multicolumn{6}{l}{\textbf{Supervision: Fully Supervised}}
        \\
        MiKASA \cite{chang2024mikasa}                   & $69.7$ & $59.4$ & $65.4$ & $64.0$ & $64.4$ \\
        ViL3DRel \cite{chen2022language_vil3drel}       & $70.2$ & $57.4$ & $62.0$ & $64.5$ & $64.4$ \\
        SceneVerse \cite{jia2024sceneverse}             & $72.5$ & $57.8$ & $56.9$ & $67.9$ & $64.9$ \\
        
        \midrule
        \multicolumn{6}{l}{\textbf{Supervision: Zero-Shot}} 
        \\
        ZSVG3D \cite{yuan2024visual_zsvg3d}             & $46.5$ & $31.7$ & $36.8$ & $40.0$ & $39.0$ \\
        VLM-Grounder \cite{xu2025vlm}                   & $\underline{55.2}$ & $\underline{39.5}$ & $\underline{45.8}$ & $\underline{49.4}$ & $\underline{48.0}$ \\
        SeeGround \cite{li2025seeground}                & $54.5$ & $38.3$ & $42.3$ & $48.2$ & $46.1$ \\
        \name{} (Ours)                                    & $\textbf{59.6}$ & $\textbf{45.4}$ & $\textbf{47.9}$ & $\textbf{54.5}$ & $\textbf{52.4}$ \\
        
        \bottomrule
    \end{tabular}
  \vspace{-0.2cm}
    \label{tab:nr3d_results}
\end{table}

\subsection{Comparative Study}

\textbf{ScanRefer.} Table \ref{tab:scanrefer_results} compares our proposed \name{} with SeeGround~\cite{li2025seeground}, a closely related zero-shot agent-based method. On the ScanRefer validation set, \name{} achieves an overall Acc@0.5 of 41.9, outperforming SeeGround by 2.5\% in this metric. Breaking down by subset: in the \textit{Unique} category, we achieve Acc@0.5 of 73.7 compared to SeeGround's 68.9 (4.8\% gain), demonstrating stronger spatial precision under stricter IoU thresholds. In the \textit{Multiple} subset with same-class distractors, \name{} obtains 31.7 Acc@0.5 versus SeeGround's 30.0, showing consistent improvements even in cluttered scenes requiring fine-grained reasoning.

\textbf{Nr3D.} Table \ref{tab:nr3d_results} presents results on the Nr3D dataset, where \name{} achieves an overall accuracy of 52.4, a 6.3\% improvement over SeeGround (46.1). Gains are consistent across difficulty and viewpoint splits: on \textit{Hard} queries, we reach 45.4 versus 38.3 (+7.1\%), indicating stronger robustness to multiple distractors, while on \textit{Easy} queries we improve to 59.6 over 54.5 (+5.1\%), suggesting better precision even with fewer confounders. For viewpoint splits, \name{} achieves 54.5 on \textit{View-Independent} compared to 48.2 (+6.3\%) and 47.9 on \textit{View-Dependent} versus 42.3 (+5.6\%), showing improvements in both geometry-driven and appearance-sensitive cases.

\textbf{Discussion.} The performance superiority of \name{} can be attributed to three key factors working in concert.

\textit{(1) Structured tool-driven agent design:} Unlike SeeGround's fixed anchor-target decomposition pipeline, our agent dynamically reasons over each query by invoking tools in a learned order. For instance, given a query like ``\texttt{the small chair on the left},'' the agent first retrieves chairs from the OLT by label, then applies spatial predicates (\texttt{smallest}, \texttt{leftmost}) to rank candidates, adapting the reasoning strategy to query complexity. This flexible decomposition reduces ambiguity when anchor or target labels are absent or ambiguous, and avoids cascading errors from misidentifying reference objects upfront.

\textit{(2) Deterministic geometric scoring:} Our agent ranks candidates using explicit geometric relations computed from OLT centers and box sizes (e.g., L2 distance, coordinate comparison). This contrasts with SeeGround's learned reranking via visual embeddings. Deterministic scoring is more stable and transparent: it guarantees high precision at stricter IoU thresholds (Acc@0.5) because spatial relationships directly encode geometric proximity without learned biases. This explains our 4.8\% gain on ScanRefer \textit{Unique} and the 6.3\% improvement on Nr3D \textit{View-Independent} queries, where viewpoint-agnostic spatial cues dominate.

\textit{(3) Agent-decided on-demand rendering:} Rather than precomputing visual attributes for all objects (as in SeeGround's attribute enrichment), our agent renders images only when: (i) visual attributes like color or material are explicitly mentioned, or (ii) geometric ambiguity arises in view-dependent queries. For the query ``\texttt{the small chair on the left},'' if geometric ranking among small chairs is ambiguous, the agent renders only those candidates' images for final visual confirmation. This selective rendering reduces cascading NLP matching errors and ensures visual cues target query-relevant objects, leading to more focused reasoning and improved Hard and Multiple subset performance.

\textit{(4) Efficient context utilization:} A practical advantage of our tool-driven design is that the agent retrieves and processes only query-relevant object metadata and images, rather than passing all scene information into a single prompt. In contrast, SeeGround's attribute enrichment pipeline predicts visual attributes for all objects in the OLT and passes the complete object list alongside rendered images into a single prompt to the VLM. This approach introduces significant token overhead, as the model must process irrelevant object information alongside the query, wasting limited context window capacity. Our selective retrieval strategy reduces the input size to the prompt, enabling more efficient processing, faster latency, and lower computational cost while maintaining or improving reasoning quality.

\subsection{Ablation Study}

\begin{table}[t]
    \centering
    \caption{Ablation study of agent tools on a held-out Nr3D subset (\mbox{$N=120$}), measured by \mbox{$Acc@0.5$}.}
    \vspace{-0.2cm}
    \begin{tabular}{l|cccc|cccc|c}
        \toprule
        \textbf{\#} & \rotatebox{90}{\textbf{Retrieval}} & \rotatebox{90}{\textbf{Distance}} & \rotatebox{90}{\textbf{Planning}} & \rotatebox{90}{\textbf{Rendering}} & \textbf{Easy} & \textbf{Hard} & \textbf{Dep} & \textbf{Indep} & \textbf{Overall} \\
        \midrule
        1 & \cmark & \xmark & \xmark & \xmark & 51.7 & 29.0 & 35.7 & 42.3 & 40.0 \\
        2 & \cmark & \cmark & \xmark & \xmark & 50.0 & \textbf{40.3} & 35.7 & \underline{50.0} & 45.0 \\
        3 & \cmark & \cmark & \cmark & \xmark & 58.6 & 35.5 & 38.1 & \textbf{51.3} & 46.7 \\
        4 & \cmark & \cmark & \xmark & \cmark & \underline{62.1} & 33.9 & \textbf{50.0} & 46.2 & \underline{47.5} \\

        \midrule

        5 & \cmark & \cmark & \cmark & \cmark & \textbf{65.5} & \underline{35.5} & \underline{47.6} & \textbf{51.3} & \textbf{50.0} \\
        \bottomrule
    \end{tabular}
    \vspace{-0.3cm}
    \label{tab:ablation_tools}
\end{table}

We studied how each tool in our agent contributes to 3DVG performance on a held-out Nr3D test subset (\mbox{$N=120$}), stratified across \textit{Easy/Hard} and \textit{View-Dependent/View-Independent} queries. We evaluate overall localization accuracy at IoU threshold 0.5 (Acc@0.5).

Table~\ref{tab:ablation_tools} incrementally enables four tools: (i) \textit{Label-based OLT Retrieval}, (ii) \textit{Geometric Distance Reasoning}, (iii) \textit{Explicit Planning}, and (iv) \textit{On-demand Visual Rendering}. Starting from retrieval-only (Overall 40.0; Easy 51.7, Hard 29.0; Dep 35.7, Indep 42.3), adding geometric distance reasoning yields a large improvement to Overall 45.0, mainly driven by gains on Hard (40.3) and Indep (50.0), indicating that deterministic spatial scoring is critical for resolving distractors. Introducing explicit planning provides a smaller but consistent gain to Overall 46.7 (Easy 58.6; Dep 38.1), suggesting that structured decomposition helps the agent select and order tool calls.

Notably, enabling on-demand rendering yields the strongest improvement on \textit{view-dependent} queries: keeping retrieval+distance fixed (no planning), turning on rendering boosts Dep from 35.7 to 50.0 (+14.3), showing that visual evidence is particularly helpful when language depends on viewpoint-sensitive attributes. In contrast, Indep slightly drops from 50.0 to 46.2, so the overall gain is more modest (45.0 $\rightarrow$ 47.5). Using the full toolset achieves the best Overall accuracy (50.0) while balancing Dep (47.6) and Indep (51.3), demonstrating that geometric reasoning, planning, and selective rendering are complementary.


\section{Conclusion}

In this paper, we presented \textbf{\name}, a novel zero-shot framework for 3D Visual Grounding on colored point clouds. The key innovation lies in a tool-driven agent design powered by LVLMs that combines deterministic geometric scoring over an Object Lookup Table with on-demand image rendering. Our system operates in a two-stage pipeline: offline 3D instance segmentation via Mask3D to construct the OLT, followed by online agent reasoning that selectively retrieves query-relevant objects and renders images only when visual evidence is needed. On the ScanRefer and Nr3D benchmarks, \name{} achieves state-of-the-art zero-shot results, with 52.4\% overall accuracy on Nr3D and 41.9\% on ScanRefer, particularly excelling in view-independent queries and dense scenes with multiple distractors.

Despite these advances, \name{} faces two primary limitations. First, processing latency remains a concern due to the computational cost of on-the-fly image rendering and LVLM inference, limiting deployment in latency-critical applications. Second, the system's performance is inherently bounded by the quality of initial 3D instance segmentation; when the underlying segmentor fails to isolate meaningful object proposals, subsequent retrieval and reasoning steps are compromised, resulting in cascading errors that cannot be recovered.

To address these limitations, we envision extending \name{} with human-in-the-loop clarification and adaptive reasoning strategies. Rather than operating fully autonomously, the agent could proactively seek user feedback when facing ambiguous candidates, enabling iterative refinement through clarification questions. Additionally, expanding the toolset to include fallback detectors and scene-aware re-segmentation heuristics could mitigate initialization failure modes, making the framework more robust to diverse scene conditions and object distributions in real-world robotic deployments.



\printbibliography





\end{document}